\documentclass[10pt,journal,compsoc]{IEEEtran}
\pdfoutput=1
\ifCLASSOPTIONcompsoc
  \usepackage[nocompress]{cite}
\else
  \usepackage{cite}
\fi
\ifCLASSINFOpdf
  \usepackage[pdftex]{graphicx}
\else
\fi

\usepackage{amssymb}
\usepackage{amsmath}
\usepackage{dsfont}
\usepackage{xcolor}
\usepackage{hyperref}
\hypersetup{
	colorlinks,
	linkcolor={red!50!black},
	citecolor={blue!50!black},
	urlcolor={blue!80!black}
}
\usepackage[ruled, vlined]{algorithm2e}
\usepackage{booktabs}
\usepackage{multirow}
\usepackage{adjustbox}
\newcommand{\etal}{\textit{et al}.}
\newcommand{\ie}{\textit{i}.\textit{e}.}
\newcommand{\eg}{\textit{e}.\textit{g}.}
\newcommand{\etc}{\textit{etc}.}

\begin{document}
%
\title{Laplacian Denoising Autoencoder}
%
%
%
%

\author{Jianbo~Jiao,~\IEEEmembership{Member,~IEEE,}
        Linchao Bao, Yunchao Wei, Shengfeng He, Honghui Shi, Rynson Lau,
        and~Thomas~S.~Huang,~\IEEEmembership{Life~Fellow,~IEEE}
\IEEEcompsocitemizethanks{\IEEEcompsocthanksitem J. Jiao is with the Department of Engineering Science, University of Oxford, Oxford, UK.\protect\\
E-mail: jianbo.jiao@eng.ox.ac.uk
\IEEEcompsocthanksitem L. Bao is with Tencent AI Lab, Shenzhen, China.
\IEEEcompsocthanksitem Y. Wei is with University of Technology Sydney, Sydney, Australia.
\IEEEcompsocthanksitem S. He is with South China University of Technology, Guangzhou, China
\IEEEcompsocthanksitem H. Shi is with University of Oregon, OR, US.
\IEEEcompsocthanksitem R. Lau is with City University of Hong Kong, HK, China.
\IEEEcompsocthanksitem T. Huang is with University of Illinois at Urbana-Champaign, IL, US.
}
}

\IEEEtitleabstractindextext{%
\begin{abstract}
While deep neural networks have been shown to perform remarkably well in many machine learning tasks, labeling a large amount of ground truth data for supervised training is usually very costly to scale. Therefore, learning robust representations with unlabeled data is critical in relieving human effort and vital for many downstream tasks. Recent advances in unsupervised and self-supervised learning approaches for visual data have benefited greatly from domain knowledge. Here we are interested in a more generic unsupervised learning framework that can be easily generalized to other domains. In this paper, we propose to learn data representations with a novel type of denoising autoencoder, where the noisy input data is generated by corrupting latent clean data in the gradient domain. This can be naturally generalized to span multiple scales with a Laplacian pyramid representation of the input data. In this way, the agent learns more robust representations that exploit the underlying data structures across multiple scales.
Experiments on several visual benchmarks demonstrate that better representations can be learned with the proposed approach, compared to its counterpart with single-scale corruption and other approaches. Furthermore, we also demonstrate that the learned representations perform well when transferring to other downstream vision tasks.
\end{abstract}

\begin{IEEEkeywords}
Laplacian, DAE, self-supervised, representation learning.
\end{IEEEkeywords}}

\maketitle

\IEEEdisplaynontitleabstractindextext

%
\IEEEpeerreviewmaketitle

\IEEEraisesectionheading{\section{Introduction}\label{sec:introduction}}

%
%
%
%
\IEEEPARstart{I}{n} recent years, deep learning has made significant improvements on machine learning tasks. However, the success of deep-based approaches relies greatly on using a large amount of human labeled data for supervision, which is usually very costly and infeasible to scale on new data. Actually, humans are exceptional experts at learning abstract knowledge from unsupervised data, \ie, without knowing the specific labels of the data. Thus, how to imitate such a human cognitive ability and effectively learn robust representations from massive sums of unlabeled data in an unsupervised manner are crucial and have been attracting interests in the literature.

Representation learning is a popular framework for unsupervised learning that aims to learn transferable representations from unlabeled data~\cite{bengio2013representation}. Although great progress has been achieved for visual data by some recent advances~\cite{zhang2016colorful,zhang2017split,pathak2016context,noroozi2016unsupervised,doersch2015unsupervised,noroozi2017representation,pathak2017learning,gidaris2018unsupervised}, the approaches are mostly designed to boost the performance of high-level recognition tasks like classification and detection~\cite{girshick2015fast,xu2019deep}. We argue that good representations should benefit multiple kinds of tasks, including both high-level recognition tasks and low-level pixel-wise prediction tasks. We, in this paper, present a novel unsupervised representation learning approach that is applicable to more generic type of data and tasks. The only assumption about the input data form is that the learned representations should incorporate the underlying data structures along some certain dimensions. For example, one would expect the representations for visual data to incorporate underlying image structures along the spatial dimension, while the representations for speech data might need to be exploited along the temporal dimension.

\begin{figure}
	\centering
	\includegraphics[width=\columnwidth]{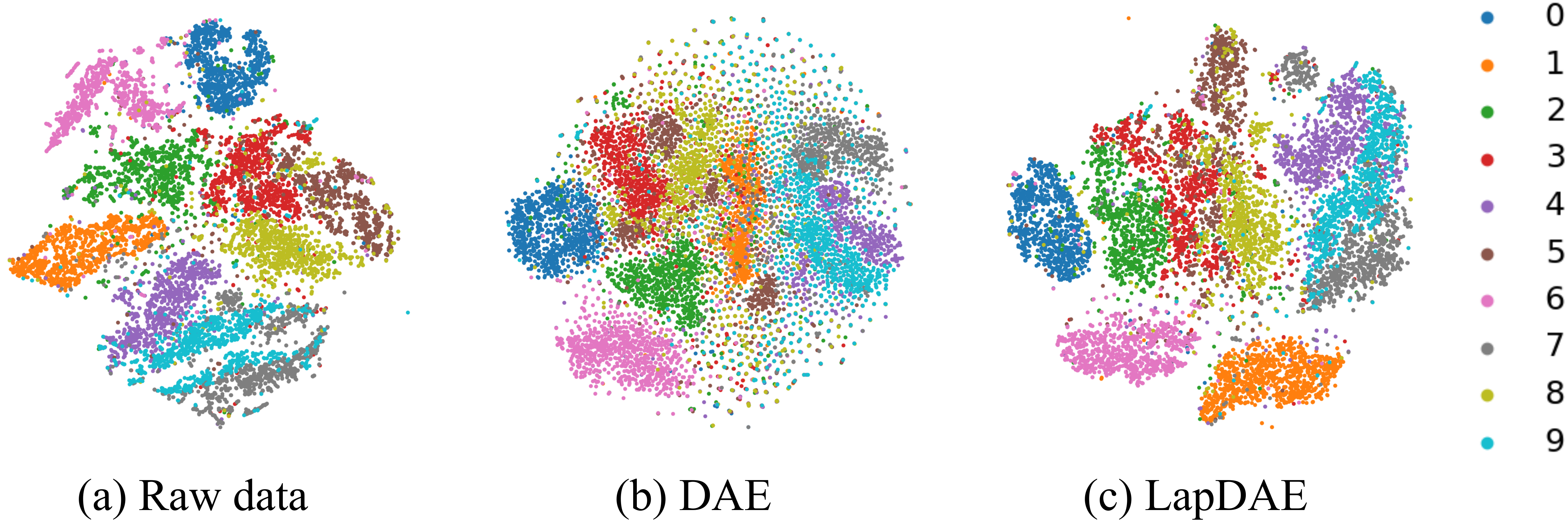}
	\caption{Illustrative visualization of the discriminative representation learning capability on the MNIST test dataset. The samples are projected to 2D domain by using the t-SNE technique~\cite{maaten2008visualizing}. (a) shows the projection from the original digits raw data. (b) is the projection from the embedding space of conventional denoising autoencoder (DAE). (c) visualizes the projected distribution from the embedding space of the proposed Laplacian denoising autoencoder (LapDAE) approach.}
	\label{fig.teas}
\end{figure}

\begin{figure*}
	\centering
	\includegraphics[width=\textwidth]{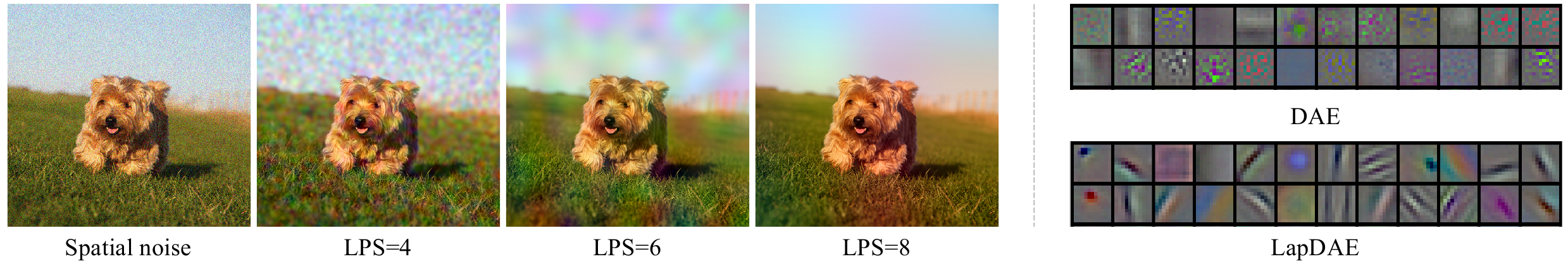}
	\caption{\textbf{Left}: illustration of our Laplacian pyramid based corruption construction strategy compared to traditional spatial corruption, where ``LPS'' indicates the Laplacian pyramid scale; \textbf{Right}: learned kernels when corruption is added in spatial domain (\emph{DAE}) and gradient domain (\emph{LapDAE}).}
	\label{fig.teas2}
\end{figure*}

Specifically, we propose to decouple the representations into different semantic levels in the Laplacian domain. A novel type of denoising autoencoder (DAE)~\cite{vincent2010stacked} is proposed to distill both high- and low-level representations accordingly. Different from the conventional DAE, where the noisy input is generated from the clean data by adding noises in the original space,  we propose to generate noisy input by corrupting the clean data in the gradient domain. By perturbing the clean data in such a manner, the corruptions are diffused into larger scales and made more difficult to remove. More importantly, the gradient domain corruption can be naturally extended to span multiple scales with a Laplacian pyramid representation of the data~\cite{burt1983laplacian}. To this end, the DAE is enforced to learn more robust and discriminative representations (Fig.~\ref{fig.teas}) that can exploit the underlying data structures across multiple scales. In addition, the proposed learning approach can easily be incorporated into other representation learning frameworks, and boosts their performance accordingly.

Our motivation is inspired by the human knowledge learning by visual perception. Instead of trying to remember every single detail, human vision focuses more on the general concept of the object/scene, which favors a combined perception of both local and non-local information~\cite{bubic2010prediction}. An example of the proposed gradient-domain corruption is illustrated in Fig.~\ref{fig.teas2}. It can be observed that compared to directly adding noise spatially, editing on different scales of the Laplacian pyramid leads to non-local random corruptions. We also show the learned kernels by the corruption in the spatial domain and that in the gradient domain on the right side of Fig.~\ref{fig.teas2}. It can be observed that more edge-sensitive and color-sensitive kernels and non-local responses are learned by the gradient domain corruption (right), in comparison to spatial corruption (left) which preferring local responses. We argue that in order for an agent to be able to recover the corruptions from different scales non-locally, it requires an understanding of the context in the presented scene.

In Fig.~\ref{fig.teas}, we illustrate the discriminative capability of our model on the MNIST~\cite{lecun1998gradient} testing set. The visualization is achieved by projecting the high-dimensional data or feature to a two-dimension space, using the t-SNE~\cite{maaten2008visualizing} technique. Compared to the raw data distribution, the embedding space of the conventional denoising autoencoder shows a better clustering ability while with some background noise. When compared to the embedding of the proposed Laplacian denoising autoencoder (LapDAE), we can observe that different categories (digits) are well discriminated from each other and with much less noise. For example, the digit \emph{5} and \emph{3} are better discriminated compared to those from the raw data and from DAE.

We demonstrate the effectiveness of the proposed unsupervised learning approach in two folds: 1) by evaluating the clustering and discriminative capability on classic benchmarks (\eg, MNIST); 2) by training on large-scale data (\eg, ImageNet~\cite{deng2009imagenet}) and transferring the learned representations to a variety of downstream vision tasks including multi-label classification, object detection, and semantic segmentation.
The main contributions of our work are summarized as follows:

\begin{itemize}
	\item We propose a new unsupervised representation learning framework , by enforcing the model to learn more context and discriminative information in the Laplacian domain.
	\item The proposed framework is trained purely based on the raw data itself and neither the data domain assumptions nor pseudo labels are necessary.
	\item Our framework is superior to the conventional DAE and achieves competitive performance on several benchmarks for representation learning.
\end{itemize}

The paper is organized as follows. We discuss related work to our approach in Section~\ref{sec:related}. The proposed LapDAE approach for self-supervised representation learning is elaborated in Section~\ref{sec:LapDAE}. Furthermore, in Section~\ref{sec:exp} we perform extensive experiments to validate the effectiveness of the proposed approach on representation learning. The transfer learning ability is also demonstrated by transferring the learned representations to several downstream tasks. Finally, the paper is concluded in Section~\ref{sec:conc} with discussion on potential future directions.

\section{Related Works}\label{sec:related}
\subsection{Autoencoders}
The conventional autoencoder (AE)~\cite{hinton2006reducing} is based on the idea of learning a mapping from high-dimension to low-dimension so that the encoded representation can be used to reconstruct the original raw input. Bengio \etal~\cite{bengio2007greedy} propose to learn the abstract representation by stacking single-layer autoencoders. Poultney \etal~\cite{poultney2007efficient} impose a sparsity prior for the latent encoded space. Furthermore, the denoising autoencoder (DAE)~\cite{vincent2010stacked} is proposed to achieve abstraction that is robust to noise and is proven to be able to learn better representations. The variational autoencoder (VAE)~\cite{kingma2013auto} aims to learn a parametric latent variable model by encouraging the latent space to satisfy a distribution.	We refer the readers to~\cite{bengio2013representation} for a broader view of autoencoder-based approaches in the literature.

\begin{figure*}
	\centering
	\includegraphics[width=0.91\textwidth]{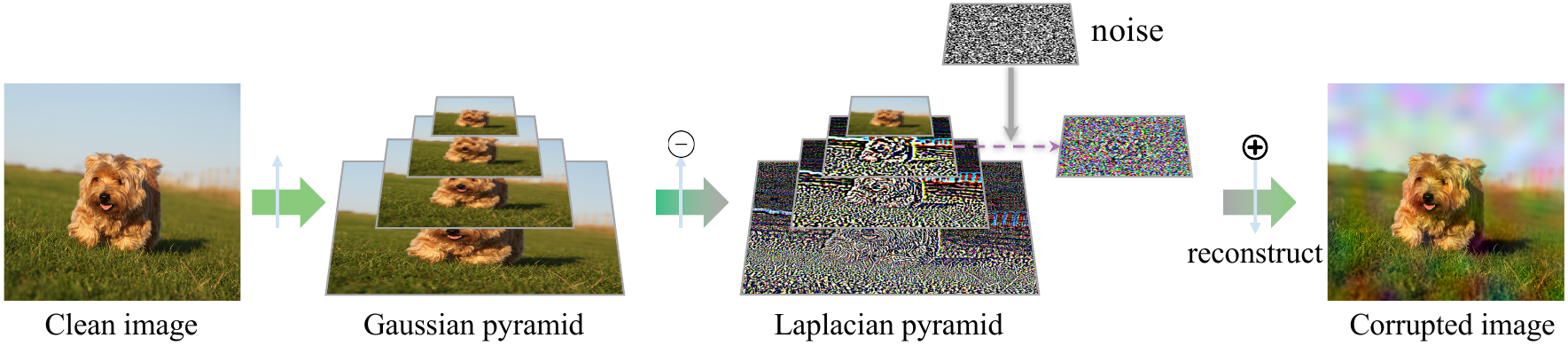}
	\caption{Illustration of the corruption with a Laplacian pyramid. A Gaussian pyramid is first constructed from the clean image, from which a Laplacian pyramid is built. After adding random corruptions (\eg, noise) to a randomly selected level (slice) in the Laplacian pyramid, the final corrupted image is reconstructed from the modified Laplacian pyramid. $\ominus$ with $\uparrow$ indicates the Laplacian pyramid construction, while $\oplus$ with $\downarrow$ indicates reconstruction.}
	\label{fig.lap}
\end{figure*}

\subsection{Representation Learning}
As a fundamental problem, representation learning has been studied for years. A comprehensive review could be referred to~\cite{bengio2013representation}. Early classical algorithms mainly focus on reconstructing the raw data by learning compressed features~\cite{hinton2006reducing,vincent2010stacked}. Some other methods instead use probabilistic models like Boltzmann machines~\cite{hinton1986learning,salakhutdinov2010efficient} and GANs~\cite{goodfellow2014generative,donahue2016adversarial}. Recently, some approaches address the problem by defining pretext tasks (termed ``self-supervised learning'') which have shown promising performance, including predicting relative spatial location/ordering of image patches~\cite{doersch2015unsupervised,noroozi2016unsupervised,noroozi2017representation}, motion in video~\cite{misra2016shuffle,pathak2017learning,wang2015unsupervised}, colorization~\cite{zhang2016colorful,zhang2017split}, predicting rotations~\cite{gidaris2018unsupervised} and transformations~\cite{zhang2019aet}, to name a few. Such pretext-task-based methods can be categorized into a different group compared to the AE/DAE-based ones. Compared to these representation learning methods, the proposed approach does not make any assumptions to the data and is more generic accordingly.

\section{Laplacian Denoising Autoencoder}\label{sec:LapDAE}
In this section, we first introduce some background information that motivates our approach. Afterwards, we elaborate the proposed Laplacian Denoising AutoEncoder (LapDAE) in detail with methodology and architecture.

\subsection{Background}
\subsubsection{Denoising Autoencoder}
In order to avoid the ``simply copy input'' that may occur in the traditional autoencoder, in \cite{vincent2010stacked} the authors introduce the denoising autoencoder (DAE), to reconstruct a ``repaired input'' from a corrupted version of it. Suppose the input is $x$, for DAE, the corrupted version $\widetilde{x}$ is mapped to a hidden representation $y=f_\theta(\widetilde{x})=s(W\widetilde{x}+b)$, by basic AE. The reconstruction from the hidden representation $y$ is $z=g_{\theta'}(y)$. The network parameters $\theta$ and $\theta'$ are learned by minimizing a reconstruction error, \eg, mean square error $L_2(x,z)=\parallel z-x\parallel_2^2$. Such a DAE is shown to learn better representation compared to the AE in its classical form~\cite{vincent2010stacked}.
\subsubsection{Laplacian Pyramid and Gradient Domain Editing}
The original Laplacian pyramid was proposed for image editing~\cite{burt1983laplacian} and can easily be generalized to other types of data where a low-pass filter is applicable. Given input data $x$, its Gaussian pyramid is composed of a set of progressively lower resolution versions of the data, denoted as $\{x^G_l\}$ where $l$ is a pyramid level. In the pyramid, the bottom level is the data itself, \ie, $x^G_0 = x$, and $x^G_{l+1}=\text{downsample}(x^G_l)$. The Laplacian pyramid $\{x^L_l\}$ is constructed by subtracting the neighboring levels in the Gaussian pyramid, $x^L_{l}=x^G_l - \text{upsample}(x^G_{l+1})$. Note that the top level of the Laplacian pyramid is the residual and the same as that in the Gaussian pyramid, $x^L_N = x^G_N$, where $N$ is the top level of the pyramid. The construction process is illustrated in Fig.~\ref{fig.lap}. Given a Laplacian pyramid, the original data can be reconstructed by recursively applying $x^G_l = x^L_l + \text{upsample}(x^G_{l+1})$ until $x^G_0$ is reached. Gradient domain editing on $x$ can be achieved by editing its Laplacian pyramid and then reconstruct the resulting $\tilde{x}$ from the modified Laplacian pyramid.

\subsection{LapDAE Methodology}
Following the denoising autoencoder framework~\cite{vincent2010stacked}, we attempt to distill the essential representations by training a convolutional network (ConvNet) to restore the clean data $x\in\mathcal{X}$ from the corrupted data $\tilde{x}\in\widetilde{\mathcal{X}}$. In contrast to a standard DAE, we generate the corrupted data $\tilde{x}$ from $x$ with the aid of a Laplacian pyramid. Specifically, we construct a Laplacian pyramid from the clean data $x$ and randomly corrupt a level of the pyramid, such that $\tilde{x}$ is reconstructed from the corrupted pyramid. Fig.~\ref{fig.lap} illustrates the process of the corruption with an example of image data. Since the corruption applied to higher levels of the pyramid affects larger spatial scales of the image (see Fig.~\ref{fig.teas2}), the randomly corrupted levels will enforce the network to learn features that can represent underlying structures across multiple scales. As known in the literature~\cite{zeiler2014visualizing}, ConvNet is inherently in favor of both local and non-local features at different levels of layers. Hence, with only local disturbances, it is difficult to capture the non-local semantic concepts. This has also been verified to some extent in the self-supervised learning methods that attempt to leverage patch-based context information~\cite{doersch2015unsupervised,noroozi2016unsupervised,noroozi2017representation}, and similarly to the non-local scheme~\cite{wang2018non} on the network design side. By adding corruptions across multiple scales, the objective is to capture both local and non-local information during the representation learning phase. Additionally, in order to incorporate diverse types of corruptions and to force the network to ``learn harder'', it is also possible to apply multiple types of corruptions to the pyramid during learning.

\begin{figure}
	\centering
	\includegraphics[width=\columnwidth]{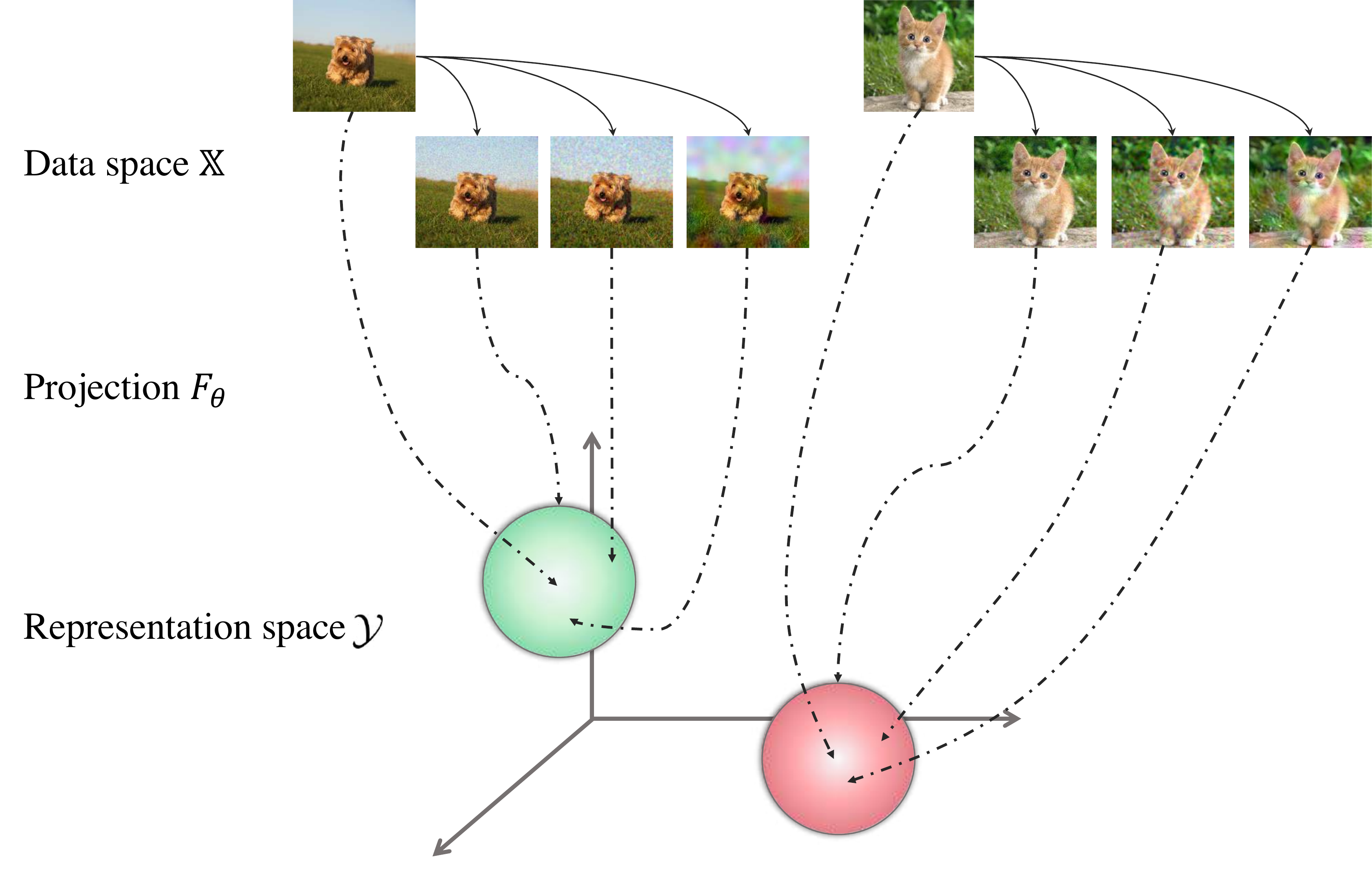}
	\caption{Illustration on semantic-invariant projection. Images (including both the original latent image and its variants in Laplacian domain) in the data space ($\mathds{X}=\mathcal{X}\cup\widetilde{\mathcal{X}}$) are projected to the representation space $\mathcal{Y}$ by a learning-based function $F_\theta$. The green and red spheres indicate different semantic-invariant representations.}
	\label{fig:proj}
\end{figure}

The assumption here is, by corrupting data in the Laplacian domain and reconstruct the original latent data from these multiple corrupted versions, the model need to understand the underlying semantic features and extract semantic-invariant representations accordingly. As illustrated in Fig.~\ref{fig:proj}, in the representation space, the original data and its Laplacian-corrupted versions are projected to a same sample sphere, where the underlying semantic-invariant representation lives. This is achieved by enforcing the reconstructions from these projected features on the same sphere to be similar. Different spheres correspond to different semantic representations, \eg, the \emph{dog} and \emph{cat} samples are projected to different representation spheres.

Denote the corrupted data as $\tilde{x}=Lap(x;\widetilde{x_l})$ (note that we use $\widetilde{x_l}$ instead of $\widetilde{x^L_l}$ for conciseness to denote a corrupted level $l$ in the Laplacian pyramid). The corrupted input space is achieved as
\begin{equation}
\mathcal{L}=\left\{Lap(x;\widetilde{x_l})_c, c\in\mathcal{C}\right\},
\end{equation}
where $\mathcal{C}$ is the corruption type set. Then the corrupted space is further mapped to a hidden (representation) space through an encoder with parameter $\theta$ by
\begin{equation}
\mathcal{Y}=F_\theta(\mathcal{L}).
\end{equation}

Differing from a DAE that uses sparse code as a hidden representation, each sample in our hidden space has its own resolution, from which we recover the reconstruction space $\mathcal{Z}=G_{\theta'}(\mathcal{Y})$ with a decoder with parameter $\theta'$. During the optimization in mini-batch, each time a training sample $x$ is presented, one or multiple versions of corruptions are constructed according to $\mathcal{C}$. Therefore each time the optimization is based on a sub-mini-batch, for which a sub-batch reconstruction objective is defined as:
\begin{equation}\label{eq.rec}
L_{\mathrm{rec}}=\underset{c \in \mathcal{C}}{\sum}\mathbb{E}_x\parallel x-z_c\parallel_2^2,
\end{equation}
where $z_c \in \mathcal{Z}$ is the reconstructed data by the network.
The learning process of the proposed LapDAE is summarized in Algorithm~\ref{algo}.

The proposed Laplacian denoising autoencoder performs data reconstruction in the Laplacian pyramid space across multiple scales. Despite simple and making no assumptions about the data and requiring no specially designed domain-specific loss functions, the proposed framework is able to learn representations competitively with existing unsupervised (as well as some self-supervised) approaches. This will be shown in the following evaluation section. Similar to some recent work which has explored withholding parts of the data (\eg, AE to remove noise; inpainting and context-encoder to drop data in spatial domain; colorization to drop data along channel direction), our LapDAE model can be considered as removing context-aware \emph{noise} along the scale direction in Laplacian domain.

Being a purely unsupervised model, this is a generic framework that can be applied to other domains in addition to visual data. The proposed framework opens a new potential direction for representation learning in another transferred domain (\eg, gradient domain), which we believe to be beneficial to the community, where current work  focuses mainly on knowledge mining in the original (spatial) domain.

\begin{algorithm}[t]
	\SetAlgoLined
	Initialize corruption set $\mathcal{C}$ and parameters $\theta,\theta'$\\
	\While{not converged}{
		\For{$x\in\mathcal{X}$}{
			Compose the Gaussian pyramid $\{x_l^G\}$\\
			Construct the Laplacian pyramid $\{x_l^L\}$ from $\{x_l^G\}$\\
			\For{$c\in\mathcal{C}$}{
				Randomly select a pyramid level $l$\\
				$\widetilde{x_l}\leftarrow$ apply corruption $c$ on level $l$\\
			}
			Reconstruct the corrupted data $\tilde{x}=Lap(x;\widetilde{x_l})$ in image domain\\
			$y=F_\theta(\tilde{x})$\\
			$z=G_{\theta'}(y)$\\
		}
		$\min \underset{x \in \mathcal{X}}{\sum} \underset{c \in \mathcal{C}}{\sum}\parallel x-z_c\parallel_2^2$
	}
	Return $\theta,\theta'$ for the LapDAE
	\caption{LapDAE Optimization}\label{algo}
\end{algorithm}

\subsection{LapDAE Architecture}\label{sec:arch}
In this section, we describe the detailed architecture of the proposed LapDAE. Specifically, we utilize a convolutional neural network (CNN) to implement the LapDAE and showcase its effectiveness on visual data. The corruption in the Laplacian domain is modeled as a Laplacian layer. Specifically, we randomly add Gaussian noise (with $\sigma=25$) to a randomly chosen level $l$ in the Laplacian pyramid. The encoder consists of several simple convolutional (\emph{conv}) layers, while the decoder is of a mirrored structure to the encoder and consists of \emph{up-conv} (also termed \emph{deconv} in some literature) layers. The model is trained with supervision from the reconstruction objective defined in Equation~(\ref{eq.rec}).

\section{Experiments}\label{sec:exp}

\subsection{Experimental Setup}
The setup of the proposed framework is described in Sec.~\ref{sec:arch}. For the basic LapDAE architecture, four $3\times3~conv$ layers are used to construct the encoder, while the decoder consists of three similar \emph{up-conv} layers. For the experiments performed on large-scale datasets (Sec.~\ref{sec:IN},~\ref{sec:Transf}), we use the AlexNet~\cite{krizhevsky2012imagenet} structure for the encoder and similarly the decoder consists of three \emph{up-conv} layers. For simplicity in this study, only one corruption type, random noise, is set in $\mathcal{C}$. The Laplacian pyramid is constructed with eight levels. The whole model is trained end-to-end by the Adam optimizer~\cite{kingma2014adam}, with the learning rate set to $10^{-4}$. The learning rate decreases at a factor of $10^{-1}$ for every 20 epochs. A batch size of 128 is used throughout the experiments.

\begin{figure*}
	\centering
	\includegraphics[width=0.96\textwidth]{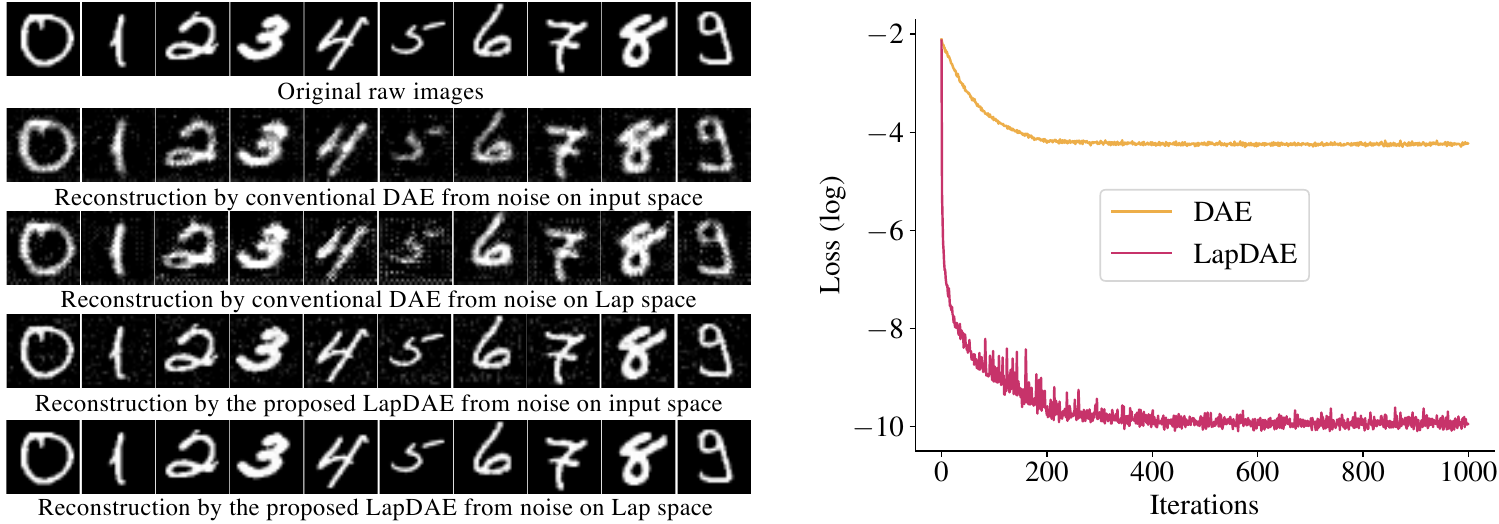}
	\caption{\textbf{Left}: An illustration of the reconstruction performance on the MNIST dataset. The original raw input images are randomly selected from the test set and are shown in the first row, while the second and last rows show the reconstructed results from conventional DAE and our LapDAE, respectively. DAE applied on Lap noise space and LapDAE on spatial noise space are also shown for reference in the third and fourth rows. \textbf{Right}: Illustration on model convergence. The horizontal axis shows the training iterations while vertical axis the training loss (in log scale).}
	\label{fig:mnist}
\end{figure*}

\subsection{Evaluation on MNIST}
The MNIST~\footnote{\url{http://yann.lecun.com/exdb/mnist/}} dataset consists of 70,000 images of handwritten digits with size of $28\times28$, in which 60,000 are used for training and the rest 10,000 for testing. Randomly selected example images from the MNIST are shown in Fig.~\ref{fig:mnist} (left). In this experiment, the input images are fed into the model at fixed size of $28\times28$ with only horizontal flipping as data augmentation during training. As the objective for our model is set as the reconstruction error (as in Equation~\ref{eq.rec}), we first illustrate the qualitative performance on the image reconstruction, shown in Fig.~\ref{fig:mnist} (left). As we can see from the reconstruction results, the conventional DAE generally reconstructs the digits but they are unclear and include some noise. In contrast, the reconstruction of our LapDAE model is evidently much clearer and includes more details, \eg, the numbers \emph{0} and \emph{4}. To better understand the reconstruction capability, we apply the conventional DAE to images corrupted with Laplacian noise, with comparison to applying our LapDAE to images where the noise is added on the input space. The results again suggest that the proposed LapDAE performs better on reconstructing context information, \ie, digits here. We also compare the convergence property of the proposed LapDAE and conventional DAE, as shown in Fig.~\ref{fig:mnist} (right). It can be observed that with the proposed LapDAE, the model converges faster and results in reaching a much more optimum level.

We perform an experiment on image clustering for both DAE and the proposed LapDAE models. The result is shown in Fig.~\ref{fig.teas}. From the results, we can see that the proposed LapDAE has a far better discriminative capability compared to the conventional DAE.

\begin{figure}
	\centering
	\includegraphics[width=\columnwidth]{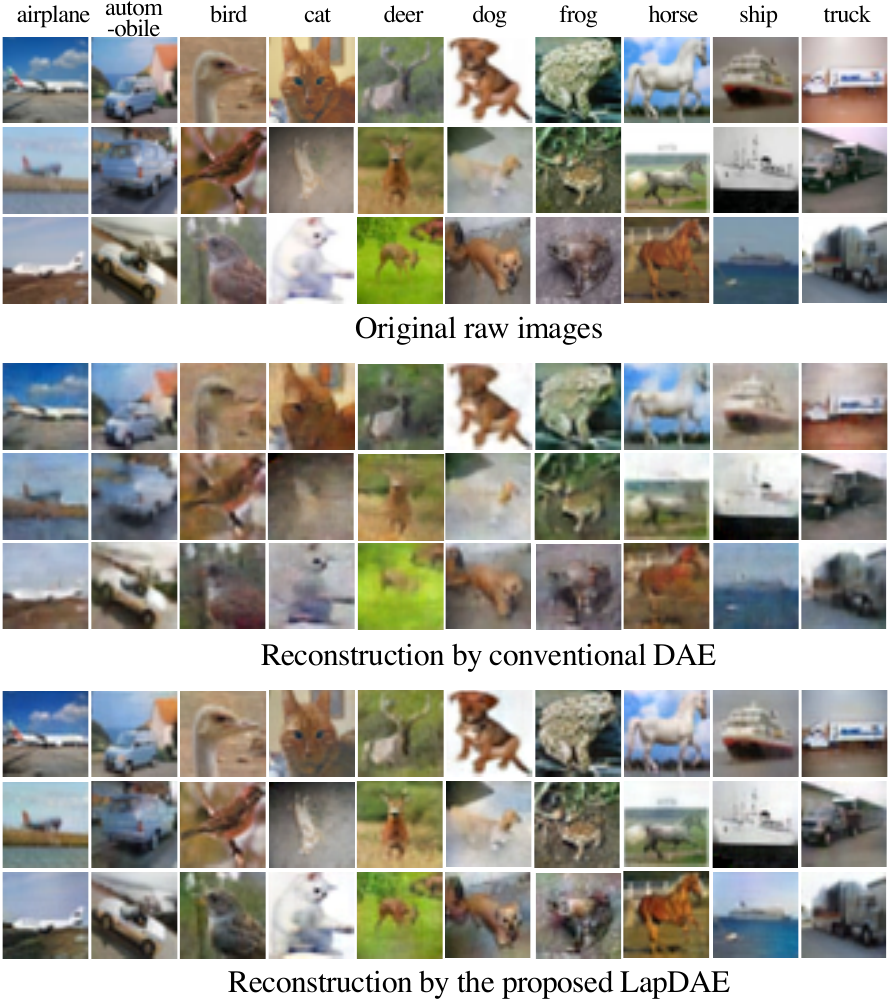}
	\caption{The reconstruction result comparison on the CIFAR-10 dataset, in which examples from each category are randomly selected for visualization.}
	\label{fig.cifar10_1}
\end{figure}

\subsection{Evaluation on CIFAR}
\begin{figure}
	\centering
	\includegraphics[width=\columnwidth]{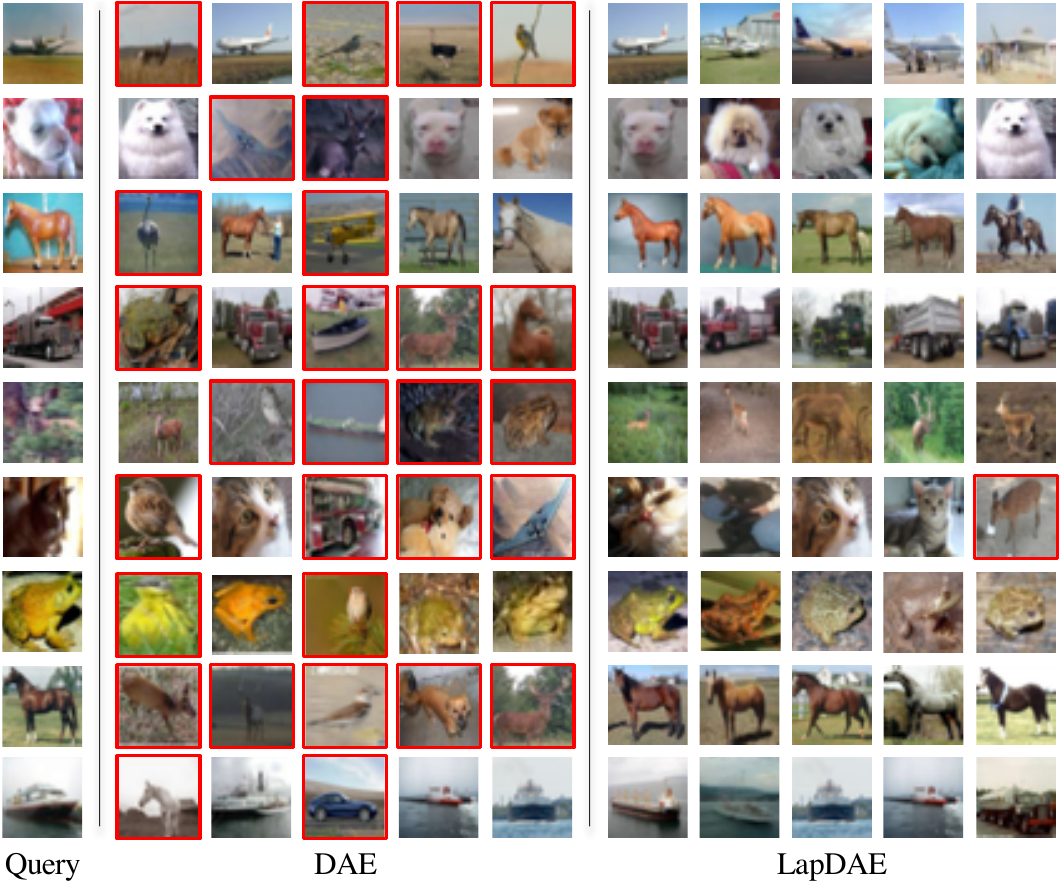}
	\caption{The image retrieval results by nearest neighbor on the CIFAR-10 dataset. Given the query on the left, the top-5 (from left to right) retrieved results of DAE (middle) and LapDAE (right) are presented, in which red ones indicate wrong category.}
	\label{fig.cifar10_2}
\end{figure}

In comparison to the MNIST dataset, the CIFAR-10~\cite{krizhevsky2009learning} dataset is composed of RGB natural images with a size of $32\times32$, covering 10 different categories of natural objects. The training set consists of 50,000 images while the testing set is 10,000. Each category includes 6,000 images. In this experiment, we first visualize the reconstructed images and compare to those reconstructed by the conventional DAE, as shown in Fig.~\ref{fig.cifar10_1}. From the results we can see that by using the proposed LapDAE, the representative context is well reconstructed, \eg, the face of the \emph{cat} and the \emph{deer} in the forest. We also explore the quality of the learned representation, by performing an image retrieval task. The retrieval is based on the similarity of the embedding space, by using the nearest neighbor scheme. Given an input query image, the feature at the bottleneck (latent space) of the model is extracted for the retrieval in the whole testing set. For this experiment, we compare with results from the conventional DAE. Fig.~\ref{fig.cifar10_2} shows example results. From the results in Fig.~\ref{fig.cifar10_2} we can observe that our LapDAE model learns a much better representation compared to the DAE. The conventional DAE tends to retrieve based on the appearance of the images, while our LapDAE focuses more on the context/semantic information, \eg, the \emph{airplane} in the first row. We attribute this to the multiple scale corruptions in the Laplacian domain. Overall, these results together with the above evaluation on the MNIST dataset can be considered as proof of concept that the proposed LapDAE is capable of capturing both low-level and high-level context information.

\subsection{Evaluation on ImageNet}\label{sec:IN}
In this section, we aim to investigate the representation learning capability of the proposed LapDAE on large-scale dataset. To achieve this goal, we perform evaluations on the ImageNet~\cite{deng2009imagenet} dataset. Specifically, we use the training set without labels from ImageNet~\cite{deng2009imagenet} to train our LapDAE model. The training set includes 1.2 million images covering 1,000 categories. Each image is first resized to $256\times256$ and randomly cropped to $227\time227$. Horizontal flipping is also applied for data augmentation.

\subsubsection{Conv1 Learned Filter Visualization}
\begin{figure}
	\centering
	\includegraphics[width=\columnwidth]{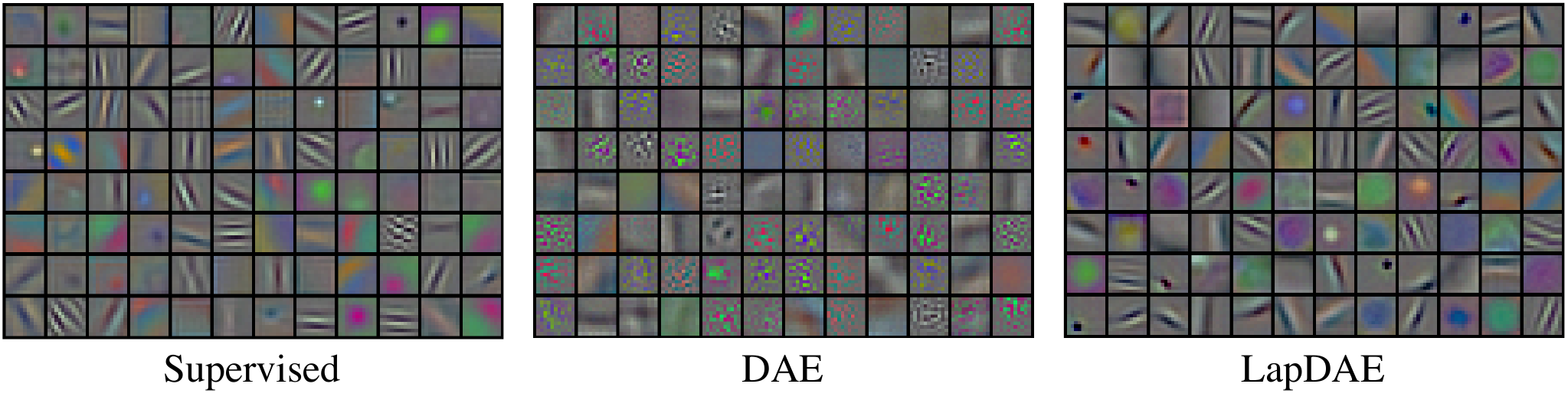}
	\caption{The learned convolutional filters (kernels) from the first layer of AlexNet (\emph{conv1}) trained on the ImageNet dataset. Left: result with fully-supervision from the labeled data;  Middle and Right: the filters learned in an unsupervised manner by DAE and our LapDAE respectively.}
	\label{fig.kernel}
\end{figure}

In Fig.~\ref{fig.kernel}, we show the comparison for the learned filters from the first layer (\ie, \emph{conv1}) of AlexNet between our approach and the fully-supervised ones. In the supervised version (the left panel), both color blobs and edge filters are learned. We can see that although not as sharp as those filters learned by the supervised setup for some blobs, our approach (the right panel) learns quite good filters including edges along different directions, edges with different frequencies, color contrast along different directions, \etc, similar to the supervised ones. Comparing with conventional DAE (the middle panel), the learned representations from our approach are much better.

\begin{table}
	\setlength{\tabcolsep}{2.5pt}
	\centering
	\caption{Top-1 accuracy on ImageNet classification with a linear classifier. The results are reported based on the ImageNet validation set and all the listed methods except \emph{ImageNet labels, Domain Adapt} are pre-trained on ImageNet without ground truth labels.}
	\label{tb.imagenet}
	\begin{tabular}{@{}lccccc@{}}
		\toprule
		Method & Conv1 & Conv2 & Conv3 & Conv4 & Conv5 \\ \midrule
		ImageNet / Places labels   & 19.3 & 36.3 & 44.2 & 48.3 & 50.5 \\ \midrule
		Random Gaussian   & 11.6 & 17.1 & 16.9 & 16.3 & 14.1 \\
		\textcolor{black}{Random rescaled~\cite{krahenbuhl2015data}}   & \textcolor{black}{17.5} & \textcolor{black}{23.0} & \textcolor{black}{24.5} & \textcolor{black}{23.2} & \textcolor{black}{20.6} \\ \midrule
		\textcolor{black}{Context~\cite{doersch2015unsupervised}}           & \textcolor{black}{16.2} & \textcolor{black}{23.3} & \textcolor{black}{30.2} & \textcolor{black}{31.7} & \textcolor{black}{29.6} \\
		Context Encoder~\cite{pathak2016context}   & 14.1 & 20.7 & 21.0 & 19.8 & 15.5 \\
		\textcolor{black}{Colorization~\cite{zhang2016colorful}}      & \textcolor{black}{12.5} & \textcolor{black}{24.5} & \textcolor{black}{30.4} & \textcolor{black}{31.5} & \textcolor{black}{30.3} \\
		\textcolor{black}{Jigsaw~\cite{noroozi2016unsupervised}}            & \textcolor{black}{18.2} & \textcolor{black}{28.8} & \textcolor{black}{34.0} & \textcolor{black}{33.9} & \textcolor{black}{27.1} \\
		BiGAN~\cite{donahue2016adversarial}             & 17.7 & 24.5 & 31.0 & 29.9 & 28.0 \\
		\textcolor{black}{Split-Brain~\cite{zhang2017split}}       & \textcolor{black}{17.7} & \textcolor{black}{29.3} & \textcolor{black}{35.4} & \textcolor{black}{35.2} & \textcolor{black}{32.8} \\
		\textcolor{black}{Counting~\cite{noroozi2017representation}}          & \textcolor{black}{18.0} & \textcolor{black}{30.6} & \textcolor{black}{34.3} & \textcolor{black}{32.5} & \textcolor{black}{25.7} \\
		RotNet~\cite{gidaris2018unsupervised} & 18.8 & 31.7 & 38.7 & 38.2 & 36.5\\
		Domain Adapt~\cite{ren-cvpr2018} & 16.5 &27.0 &30.5 & 30.1 &26.5\\ 
		Instance~\cite{wu2018unsupervised} & 16.8 &26.5 &31.8 & 34.1 &35.6\\
		AND~\cite{huang2018and} & 15.6 &27.0 &35.9 & 39.7 &37.9\\ \midrule
		AET-project~\cite{zhang2019aet} & 19.2 & 32.8 & 40.6 & 39.7 & 37.7 \\
		DAE   & 12.5 & 18.5 & 21.8 & 20.4 & 14.8 \\
		DAE + Trans   & 17.6  & 31.8 & 39.2 & 37.4 & 34.5 \\
		(Ours) LapDAE  &18.4 & 27.4 & 29.9 & 27.0 & 22.7 \\
		(Ours) LapDAE + Trans &\textbf{19.3} &\textbf{33.2} &\textbf{43.2} &\textbf{41.1} &\textbf{39.6}\\
		\bottomrule
	\end{tabular}
\end{table}

\subsubsection{Controlled Classification}
Here we quantitatively evaluate our learned representations on the ImageNet classification task~\cite{deng2009imagenet}. Following the experimental settings in~\cite{zhang2016colorful}, we freeze the pre-trained weights of our model and train a linear classifier on the top of each \emph{conv} layer, to perform the 1000-category classification task. In order to have approximately the same dimensions across different layers, the feature maps of each layer is interpolated to have around 9000 elements.
Table~\ref{tb.imagenet} shows the evaluation results. Several state-of-the-art self-supervised representation learning methods~\cite{zhang2016colorful,zhang2017split,pathak2016context,doersch2015unsupervised,noroozi2016unsupervised,donahue2016adversarial,noroozi2017representation,gidaris2018unsupervised,ren-cvpr2018,wu2018unsupervised,zhang2019aet,huang2018and} are included for comparison. Since the proposed approach can be easily integrated into other representation learning frameworks, we also present the performance of our LapDAE combined with the task of predicting transformations (\emph{LapDAE+Trans}). Specifically, we base on the AET framework~\cite{zhang2019aet} while reasoning the transformation between the original image and a transformed one corrupted by our LapDAE.

From the results, we observe that the proposed method with the Laplacian pyramid largely improves the performance compared to its counterpart without the Laplacian pyramid, especially for the lower convolutional layers. This is consistent with the above visualization and analysis of the \emph{conv1} layer, where the filter kernels have more representative power in the proposed LapDAE.
When incorporating with the transformation prediction task, we can see that the performance is further boosted by a large margin. Even when compared to the \emph{AET-project} method, our approach performs much better, which again validates the effectiveness of the proposed LapDAE.

\subsection{Transfer Learning Analysis}\label{sec:Transf}

\begin{table}
	\setlength{\tabcolsep}{2.5pt}
	\centering
	\caption{Top-1 accuracy on Places classification with a linear classifier. The results are reported based on the Places validation set and all the listed methods except \emph{Places labels, Domain Adapt} are pre-trained on ImageNet without ground truth labels.}
	\label{tb.places}
	\begin{tabular}{@{}lccccc@{}}
		\toprule
		Method & Conv1 & Conv2 & Conv3 & Conv4 & Conv5 \\ \midrule
		Places labels   & 22.1 & 35.1 & 40.2 & 43.3 & 44.6 \\ \midrule
		Random Gaussian   & 15.7 & 20.3 & 19.8 & 19.1 & 17.5 \\
		\textcolor{black}{Random rescaled~\cite{krahenbuhl2015data}}   & \textcolor{black}{21.4} & \textcolor{black}{26.2} & \textcolor{black}{27.1} & \textcolor{black}{26.1} & \textcolor{black}{24.0} \\ \midrule
		\textcolor{black}{Context~\cite{doersch2015unsupervised}}           & \textcolor{black}{19.7} & \textcolor{black}{26.7} & \textcolor{black}{31.9} & \textcolor{black}{32.7} & \textcolor{black}{30.9} \\
		Context Encoder~\cite{pathak2016context}   & 18.2 & 23.2 & 23.4 & 21.9 & 18.4 \\
		\textcolor{black}{Colorization~\cite{zhang2016colorful}}      & \textcolor{black}{16.0} & \textcolor{black}{25.7} & \textcolor{black}{29.6} & \textcolor{black}{30.3} & \textcolor{black}{29.7} \\
		\textcolor{black}{Jigsaw~\cite{noroozi2016unsupervised}}            & \underline{23.0} & \textcolor{black}{31.9} & \textcolor{black}{35.0} & \textcolor{black}{34.2} & \textcolor{black}{29.3} \\
		BiGAN~\cite{donahue2016adversarial}             & 22.0 & 28.7 & 31.8 & 31.3 & 29.7 \\
		\textcolor{black}{Split-Brain~\cite{zhang2017split}}       & \textcolor{black}{21.3} & \textcolor{black}{30.7} & \textcolor{black}{34.0} & \textcolor{black}{34.1} & \textcolor{black}{32.5} \\
		\textcolor{black}{Counting~\cite{noroozi2017representation}}          & \textbf{23.3} & \textbf{33.9} & \textcolor{black}{36.3} & \textcolor{black}{34.7} & \textcolor{black}{29.6} \\
		RotNet~\cite{gidaris2018unsupervised} & 21.5 & 31.0 & 35.1 & 34.6 & 33.7\\
		Instance~\cite{wu2018unsupervised} & 18.8 & 24.3 & 31.9 & 34.5 & 33.6\\
		\midrule
		AET-project~\cite{zhang2019aet} & 22.1 & 32.9 & \underline{37.1} & \underline{36.2} & \underline{34.7} \\
		DAE   & 15.9 & 22.6 & 24.2 & 22.1 & 18.1 \\
		DAE + Trans   & 21.6  & 31.7 & 35.7 & 34.1 & 32.8 \\
		(Ours) LapDAE  & 21.0 & 30.9 & 31.6 & 29.2 & 26.1 \\
		(Ours) LapDAE + Trans &{22.2} &\underline{33.8} &\textbf{38.2} &\textbf{37.3} &\textbf{36.1}\\
		\bottomrule
	\end{tabular}
\end{table}

\subsubsection{Performance on Places}
In addition to the controlled classification on ImageNet, here we evaluate the representation learning capability by transfer learning on the Places dataset~\cite{zhou2014learning}. We use the same experimental settings as the ImageNet experiment and the result is shown in Table~\ref{tb.places}. Differing from the ImageNet experiments, the classifier trained on top of different layers is a 205-way logistic regression layer and which is then trained with the Places labels. From the result, we can see that the proposed LapDAE performs better than its counterparts and outperforms other methods when incorporating with transformation prediction.

\subsubsection{Performance on \textsc{Pascal} VOC}
Furthermore, we perform a transfer learning evaluation on the \textsc{Pascal} VOC dataset~\cite{everingham2010pascal}, for more vision tasks of multi-label classification, detection on VOC 2007, and semantic segmentation on VOC 2012. The learned weights of our model trained on ImageNet are transferred to a standard AlexNet for the evaluation. We then fine-tune the model on the \textsc{Pascal} VOC trainval set and test on the test set. Note that we do not apply any ``magic'' techniques such as weights rescaling~\cite{krahenbuhl2015data}. For the classification task, we use the same network architecture as in the ImageNet evaluation, while for the detection and semantic segmentation tasks we use the publicly available frameworks of Fast R-CNN~\cite{girshick2015fast} and FCN~\cite{long2015fully}, respectively, following the same setups for other state-of-the-art methods~\eg,
~\cite{doersch2015unsupervised,zhang2016colorful}. The results in Table~\ref{tb.pascal} show that the learned visual representations by the proposed LapDAE exhibit good performances when transferred to other vision datasets or tasks, and performs favorably against the other state-of-the-art methods. For the classification task of \textsc{Pascal} VOC, the proposed LapDAE outperforms all other generic unsupervised learning methods like DAE and GAN. More impressively, its segmentation performance on \textsc{Pascal} VOC surpasses most of the representation learning methods including those self-supervised learning approaches with specifically designed pretext tasks. Comparison between the proposed framework and its counterpart DAE suggests that the improvement is partly due to the Laplacian pyramid. When incorporating the transformation prediction task (\emph{LapDAE+Trans}), our approach on \textsc{Pascal} VOC transfer learning is further improved and achieves state-of-the-art performance.

\begin{table}
	\centering
	\small
	\setlength{\tabcolsep}{2.5pt}
	\caption{Comparison with state-of-the-art representation learning methods on \textsc{Pascal} VOC vision tasks of classification, detection on 2007, and semantic segmentation on 2012. For classification we also compare a setup that fixes the layers before \emph{conv5} and only train the \emph{fc6-8}. The learned weights from unlabeled ImageNet are transferred for the new tasks except the \emph{ImageNet labels}.}
	\label{tb.pascal}
	\begin{adjustbox}{max width=\columnwidth}
		\begin{tabular}{@{}lcccc@{}}
			\toprule
			\multirow{3}{*}{Method} & \multicolumn{2}{c}{ \begin{tabular}[c]{@{}c@{}}Classification\\ (\%mAP)\end{tabular}} & \begin{tabular}[c]{@{}c@{}}Detection \\(\%mAP)\end{tabular} & \begin{tabular}[c]{@{}c@{}}Segmentation \\(\%mIoU)\end{tabular} \\ \cmidrule(r){2-5}
			&fc6-8 & all & all & all \\ \midrule
			ImageNet labels & 78.9 & 79.9 & 56.8 & 48.0 \\ \midrule
			Random Gaussian & - & 53.3 & 43.4 & 19.8 \\
			{Random rescaled~\cite{krahenbuhl2015data}} & {39.2} & {56.6} & 45.6 & {32.6} \\
			\midrule
			Context~\cite{doersch2015unsupervised} & - & 55.3 & 45.7 & - \\
			Context Encoder~\cite{pathak2016context} & 34.6 & 56.5 & 44.5 & 29.7 \\
			{Colorization~\cite{zhang2016colorful}} & 61.5 & 65.5 & 46.9 & 35.6 \\
			Counting~\cite{noroozi2017representation} & - & 67.7 & 51.4 & 36.6 \\
			GAN~\cite{goodfellow2014generative} & 40.5 & 56.4 & - & -\\
			{BiGAN~\cite{donahue2016adversarial}} & {52.3} & {60.1} & {46.9} & {34.9} \\
			RotNet~\cite{gidaris2018unsupervised} & 70.9 & 73.0 & 54.4 & 39.1 \\
			\midrule
			AET-project~\cite{zhang2019aet} & 70.5 & 73.1 & 54.2 & 39.3 \\
			DAE & 37.0 & 54.6 & 43.4 & 29.1\\
			DAE + Trans & 66.7 & 70.1 & 51.0 & 36.8\\
			(Ours) LapDAE & 50.6 & 59.0 & 45.6 & {38.3}\\
			(Ours) LapDAE + Trans & \textbf{71.4} & \textbf{74.2}& \textbf{55.2}& \textbf{41.1} \\
			\bottomrule
		\end{tabular}
	\end{adjustbox}
\end{table}

\section{Discussion and Conclusion}\label{sec:conc}
In this paper we introduced a novel type of denoising autoencoder for unsupervised representation learning.
In contrast to conventional DAE, the corrupted data input to the proposed DAE is produced with the aid of a Laplacian pyramid.
By adding corruptions to randomly chosen levels in a Laplacian pyramid, the resulting data corruptions span multiple scales across the original data space. From this, the model is forced to learn to represent underlying data structures across multiple scales. The proposed learning framework ensures that the agent learns better representations when compared to a conventional DAE and other self-supervised learning methods.
Through extensive experimental evaluations, we demonstrated the effectiveness of the proposed LapDAE on representation learning without external supervisions. The transfer learning ability is also validated by several visual tasks.

While in this paper we showcase the effectiveness of the proposed method for learning transferable representations on vision tasks, it would be interesting to see how it performs with other types of data.
Another interesting direction worth further investigation is including additional constraints to regularize the learning procedure, for instance, by introducing contrastive loss to maximize the distance between different semantic spheres while minimize the distance between samples belonging to the same semantic sphere.
The core idea of performing both local and non-local learning that is consistent with the hierarchical nature of ConvNets, is the Laplacian pyramid space, which we believe to be a promising direction for future research.


%


\ifCLASSOPTIONcompsoc
\else
\fi


\ifCLASSOPTIONcaptionsoff
  \newpage
\fi



\bibliographystyle{IEEEtran}
\bibliography{cites}

\begin{thebibliography}{10}
\providecommand{\url}[1]{#1}
\csname url@samestyle\endcsname
\providecommand{\newblock}{\relax}
\providecommand{\bibinfo}[2]{#2}
\providecommand{\BIBentrySTDinterwordspacing}{\spaceskip=0pt\relax}
\providecommand{\BIBentryALTinterwordstretchfactor}{4}
\providecommand{\BIBentryALTinterwordspacing}{\spaceskip=\fontdimen2\font plus
\BIBentryALTinterwordstretchfactor\fontdimen3\font minus
  \fontdimen4\font\relax}
\providecommand{\BIBforeignlanguage}[2]{{%
\expandafter\ifx\csname l@#1\endcsname\relax
\typeout{** WARNING: IEEEtran.bst: No hyphenation pattern has been}%
\typeout{** loaded for the language `#1'. Using the pattern for}%
\typeout{** the default language instead.}%
\else
\language=\csname l@#1\endcsname
\fi
#2}}
\providecommand{\BIBdecl}{\relax}
\BIBdecl

\bibitem{bengio2013representation}
Y.~Bengio, A.~Courville, and P.~Vincent, ``Representation learning: A review
  and new perspectives,'' \emph{IEEE Transactions on Pattern Analysis and
  Machine Intelligence}, vol.~35, no.~8, pp. 1798--1828, 2013.

\bibitem{zhang2016colorful}
R.~Zhang, P.~Isola, and A.~A. Efros, ``Colorful image colorization,'' in
  \emph{Proceedings of European Conference on Computer Vision (ECCV)}, 2016.

\bibitem{zhang2017split}
------, ``Split-brain autoencoders: Unsupervised learning by cross-channel
  prediction,'' in \emph{Proceedings of the IEEE Conference on Computer Vision
  and Pattern Recognition (CVPR)}, 2017.

\bibitem{pathak2016context}
D.~Pathak, P.~Krahenbuhl, J.~Donahue, T.~Darrell, and A.~A. Efros, ``Context
  encoders: Feature learning by inpainting,'' in \emph{Proceedings of the IEEE
  Conference on Computer Vision and Pattern Recognition (CVPR)}, 2016.

\bibitem{noroozi2016unsupervised}
M.~Noroozi and P.~Favaro, ``Unsupervised learning of visual representations by
  solving jigsaw puzzles,'' in \emph{Proceedings of European Conference on
  Computer Vision (ECCV)}, 2016.

\bibitem{doersch2015unsupervised}
C.~Doersch, A.~Gupta, and A.~A. Efros, ``Unsupervised visual representation
  learning by context prediction,'' in \emph{Proceedings of the IEEE
  International Conference on Computer Vision (ICCV)}, 2015.

\bibitem{noroozi2017representation}
M.~Noroozi, H.~Pirsiavash, and P.~Favaro, ``Representation learning by learning
  to count,'' in \emph{Proceedings of the IEEE International Conference on
  Computer Vision (ICCV)}, 2017.

\bibitem{pathak2017learning}
D.~Pathak, R.~Girshick, P.~Doll{\'a}r, T.~Darrell, and B.~Hariharan, ``Learning
  features by watching objects move,'' in \emph{Proceedings of the IEEE
  Conference on Computer Vision and Pattern Recognition (CVPR)}, 2017.

\bibitem{gidaris2018unsupervised}
S.~Gidaris, P.~Singh, and N.~Komodakis, ``Unsupervised representation learning
  by predicting image rotations,'' \emph{arXiv preprint arXiv:1803.07728},
  2018.

\bibitem{girshick2015fast}
R.~Girshick, ``Fast r-cnn,'' in \emph{Proceedings of the IEEE international
  conference on computer vision (ICCV)}, 2015.

\bibitem{xu2019deep}
H.~Xu, X.~Lv, X.~Wang, Z.~Ren, N.~Bodla, and R.~Chellappa, ``Deep regionlets:
  Blended representation and deep learning for generic object detection,''
  \emph{IEEE Transactions on Pattern Analysis and Machine Intelligence}, 2019.

\bibitem{maaten2008visualizing}
L.~v.~d. Maaten and G.~Hinton, ``Visualizing data using t-sne,'' \emph{Journal
  of machine learning research}, vol.~9, no. Nov, pp. 2579--2605, 2008.

\bibitem{vincent2010stacked}
P.~Vincent, H.~Larochelle, I.~Lajoie, Y.~Bengio, and P.-A. Manzagol, ``Stacked
  denoising autoencoders: Learning useful representations in a deep network
  with a local denoising criterion,'' \emph{Journal of Machine Learning
  Research}, vol.~11, no. Dec, pp. 3371--3408, 2010.

\bibitem{burt1983laplacian}
P.~Burt and E.~Adelson, ``The laplacian pyramid as a compact image code,''
  \emph{IEEE Transactions on communications}, vol.~31, no.~4, pp. 532--540,
  1983.

\bibitem{bubic2010prediction}
A.~Bubi{\'c}, D.~Y. von Cramon, and R.~I. Schubotz, ``Prediction, cognition and
  the brain,'' \emph{Frontiers in Human Neuroscience}, vol.~4, no.~25, 2010.

\bibitem{lecun1998gradient}
Y.~LeCun, L.~Bottou, Y.~Bengio, P.~Haffner \emph{et~al.}, ``Gradient-based
  learning applied to document recognition,'' \emph{Proceedings of the IEEE},
  vol.~86, no.~11, pp. 2278--2324, 1998.

\bibitem{deng2009imagenet}
J.~Deng, W.~Dong, R.~Socher, L.-J. Li, K.~Li, and L.~Fei-Fei, ``Imagenet: A
  large-scale hierarchical image database,'' in \emph{Proceedings of the IEEE
  Conference on Computer Vision and Pattern Recognition (CVPR)}, 2009.

\bibitem{hinton2006reducing}
G.~E. Hinton and R.~R. Salakhutdinov, ``Reducing the dimensionality of data
  with neural networks,'' \emph{science}, vol. 313, no. 5786, pp. 504--507,
  2006.

\bibitem{bengio2007greedy}
Y.~Bengio, P.~Lamblin, D.~Popovici, and H.~Larochelle, ``Greedy layer-wise
  training of deep networks,'' in \emph{Advances in neural information
  processing systems (NeurIPS)}, 2007.

\bibitem{poultney2007efficient}
C.~Poultney, S.~Chopra, Y.~L. Cun \emph{et~al.}, ``Efficient learning of sparse
  representations with an energy-based model,'' in \emph{Advances in neural
  information processing systems (NeurIPS)}, 2007.

\bibitem{kingma2013auto}
D.~P. Kingma and M.~Welling, ``Auto-encoding variational bayes,'' \emph{arXiv
  preprint arXiv:1312.6114}, 2013.

\bibitem{hinton1986learning}
G.~E. Hinton and T.~J. Sejnowski, ``Learning and releaming in boltzmann
  machines,'' \emph{Parallel distributed processing: Explorations in the
  microstructure of cognition}, vol.~1, no. 282-317, p.~2, 1986.

\bibitem{salakhutdinov2010efficient}
R.~Salakhutdinov and H.~Larochelle, ``Efficient learning of deep boltzmann
  machines,'' in \emph{Proceedings of the Thirteenth International Conference
  on Artificial Intelligence and Statistics}, 2010.

\bibitem{goodfellow2014generative}
I.~Goodfellow, J.~Pouget-Abadie, M.~Mirza, B.~Xu, D.~Warde-Farley, S.~Ozair,
  A.~Courville, and Y.~Bengio, ``Generative adversarial nets,'' in
  \emph{Advances in neural information processing systems (NeurIPS)}, 2014.

\bibitem{donahue2016adversarial}
J.~Donahue, P.~Kr{\"a}henb{\"u}hl, and T.~Darrell, ``Adversarial feature
  learning,'' \emph{arXiv preprint arXiv:1605.09782}, 2016.

\bibitem{misra2016shuffle}
I.~Misra, C.~L. Zitnick, and M.~Hebert, ``Shuffle and learn: unsupervised
  learning using temporal order verification,'' in \emph{Proceedings of
  European Conference on Computer Vision (ECCV)}, 2016.

\bibitem{wang2015unsupervised}
X.~Wang and A.~Gupta, ``Unsupervised learning of visual representations using
  videos,'' in \emph{Proceedings of the IEEE International Conference on
  Computer Vision (ICCV)}, 2015.

\bibitem{zhang2019aet}
L.~Zhang, G.-J. Qi, L.~Wang, and J.~Luo, ``Aet vs. aed: Unsupervised
  representation learning by auto-encoding transformations rather than data,''
  in \emph{Proceedings of the IEEE Conference on Computer Vision and Pattern
  Recognition (CVPR)}, 2019.

\bibitem{zeiler2014visualizing}
M.~D. Zeiler and R.~Fergus, ``Visualizing and understanding convolutional
  networks,'' in \emph{Proceedings of European Conference on Computer Vision
  (ECCV)}, 2014.

\bibitem{wang2018non}
X.~Wang, R.~Girshick, A.~Gupta, and K.~He, ``Non-local neural networks,'' in
  \emph{Proceedings of the IEEE Conference on Computer Vision and Pattern
  Recognition (CVPR)}, 2018.

\bibitem{krizhevsky2012imagenet}
A.~Krizhevsky, I.~Sutskever, and G.~E. Hinton, ``Imagenet classification with
  deep convolutional neural networks,'' in \emph{Advances in neural information
  processing systems (NeurIPS)}, 2012.

\bibitem{kingma2014adam}
D.~P. Kingma and J.~Ba, ``Adam: A method for stochastic optimization,''
  \emph{arXiv preprint arXiv:1412.6980}, 2014.

\bibitem{krizhevsky2009learning}
A.~Krizhevsky, ``Learning multiple layers of features from tiny images,''
  Citeseer, Tech. Rep., 2009.

\bibitem{krahenbuhl2015data}
P.~Kr{\"a}henb{\"u}hl, C.~Doersch, J.~Donahue, and T.~Darrell, ``Data-dependent
  initializations of convolutional neural networks,'' \emph{arXiv preprint
  arXiv:1511.06856}, 2015.

\bibitem{ren-cvpr2018}
Z.~Ren and Y.~J. Lee, ``Cross-domain self-supervised multi-task feature
  learning using synthetic imagery,'' in \emph{Proceedings of the IEEE
  Conference on Computer Vision and Pattern Recognition (CVPR)}, 2018.

\bibitem{wu2018unsupervised}
Z.~Wu, Y.~Xiong, S.~X. Yu, and D.~Lin, ``Unsupervised feature learning via
  non-parametric instance discrimination,'' in \emph{Proceedings of the IEEE
  Conference on Computer Vision and Pattern Recognition (CVPR)}, 2018.

\bibitem{huang2018and}
J.~Huang, Q.~Dong, S.~Gong, and X.~Zhu, ``Unsupervised deep learning by
  neighbourhood discovery,'' in \emph{International Conference on Machine
  Learning (ICML)}, 2019.

\bibitem{zhou2014learning}
B.~Zhou, A.~Lapedriza, J.~Xiao, A.~Torralba, and A.~Oliva, ``Learning deep
  features for scene recognition using places database,'' in \emph{Advances in
  neural information processing systems (NeurIPS)}, 2014.

\bibitem{everingham2010pascal}
M.~Everingham, L.~Van~Gool, C.~K. Williams, J.~Winn, and A.~Zisserman, ``The
  pascal visual object classes (voc) challenge,'' \emph{International journal
  of computer vision}, vol.~88, no.~2, pp. 303--338, 2010.

\bibitem{long2015fully}
J.~Long, E.~Shelhamer, and T.~Darrell, ``Fully convolutional networks for
  semantic segmentation,'' in \emph{Proceedings of the IEEE Conference on
  Computer Vision and Pattern Recognition (CVPR)}, 2015.

\end{thebibliography}
%



\end{document}